\documentclass[conference]{IEEEtran}
\IEEEoverridecommandlockouts

\usepackage{cite}
\usepackage{amsmath,amssymb,amsfonts}
\usepackage{pifont}
\newcommand{\cmark}{\ding{51}}%
\newcommand{\xmark}{\ding{55}}%
\usepackage{algorithmic}
\usepackage{algorithm}
\usepackage{graphicx}
\usepackage{textcomp}
\usepackage{mathtools}
\usepackage{tabularx}
\usepackage{float}
\usepackage{subfigure}
\usepackage{etoolbox}
\makeatletter
\patchcmd{\@makecaption}
  {\scshape}
  {}
  {}
  {}
\makeatother
\usepackage{multirow}
\newtheorem{defn}{Definition}[section]
\newcommand{\abs}[1]{\left|#1\right|}

\usepackage{xcolor}
\def\BibTeX{{\rm B\kern-.05em{\sc i\kern-.025em b}\kern-.08em
    T\kern-.1667em\lower.7ex\hbox{E}\kern-.125emX}}

\begin{document}

\title{Representation range needs for 16-bit neural \\ network training}

\author{
\IEEEauthorblockN{1\textsuperscript{st} Valentina Popescu}
\IEEEauthorblockA{
\textit{Cerebras Systems}\\
valentina@cerebras.net}
\and
\IEEEauthorblockN{1\textsuperscript{st} Abhinav Venigalla}
\IEEEauthorblockA{
\textit{ }\\
abhinav.venigalla@gmail.com}
\footnotesize \textsuperscript{*}{work done while at Cerebras Systems}
\and
\IEEEauthorblockN{2\textsuperscript{nd} Di Wu}
\IEEEauthorblockA{
\textit{University of Wisconsin--Madison}\\
di.wu@ece.wisc.edu}
\footnotesize \textsuperscript{*}{work done while at Cerebras Systems}
\and
\IEEEauthorblockN{3\textsuperscript{rd} Robert Schreiber}
\IEEEauthorblockA{
\textit{Cerebras Systems}\\
rob.schreiber@cerebras.net}
}

\maketitle

\begin{abstract}
Deep learning has grown rapidly thanks to its state-of-the-art performance across a wide range of real-world applications.
While neural networks have been trained using IEEE-754 \texttt{binary32} arithmetic, the rapid growth of computational demands in deep learning has boosted interest in faster, low precision training. 
Mixed-precision training that combines IEEE-754 \texttt{binary16} with IEEE-754 \texttt{binary32} has been tried, and other $16$-bit formats, for example Google's \texttt{bfloat16}, have become popular.

In floating-point arithmetic there is a tradeoff between precision and representation range as the number of exponent bits changes;  \emph{denormal} numbers extend the representation range. This raises questions of how much exponent range is needed, of whether there is a format between \texttt{binary16} ($5$ exponent bits) and \texttt{bfloat16} ($8$ exponent bits) that works better than either of them, and whether or not denormals are necessary.

In the current paper we study the need for denormal numbers for mixed-precision training, and we propose a $1/6/9$ format, i.e., $6$-bit exponent and $9$-bit explicit mantissa, that offers a better range-precision tradeoff. We show that $1/6/9$ mixed-precision training is able to speed up training on hardware that incurs a performance slowdown on denormal operations or eliminates the need for denormal numbers altogether. 
And, for a number of fully connected and convolutional neural networks in computer vision and natural language processing, $1/6/9$ achieves numerical parity to standard mixed-precision.

\end{abstract}

\medskip

\section{Introduction}
Larger and more capable neural networks demand ever more computation to be trained. Industry has responded with specialized processors that can speed up machine arithmetic.  Most deep learning training is done using GPUs and/or CPUs, which support 
IEEE-754 \emph{binary32} (\emph{single}-precision) and (for GPUs) \emph{binary16} (\emph{half}-precision). 

Arithmetic and load/store instructions with $16$-bit numbers can be significantly faster than arithmetic with longer $32$-bit numbers.
Since $32$-bit is expensive for large workloads, industry and the research community have tried to use mixed-precision methods that perform some of the work in $16$-bit and the rest in $32$-bit. 
Training neural networks using $16$-bit formats is challenging because the magnitudes and the range of magnitudes of the network tensors vary greatly from layer to layer, and can sometimes shift as training progresses.

In this effort, \emph{binary16} has not proved to be an unqualified success, mainly because its $5$ exponent bits offer a narrow representation range of representable values, even when denormalized floating point numbers (or \emph{denormals} --- see a formal definition in Section~\ref{sec:fp-arith}), which add three orders of magnitude to the representation range, are used.  
Therefore, different allocations of the $15$ bits beyond the sign bit, to allow greater representation range, have been tried.
Most notably, Google introduced the \emph{bfloat16} format, with $8$ exponent bits, 
which until recently was only available on TPU clouds, but is now also included in Nvidia's latest architecture, Ampere.
Experiments have shown that both \emph{binary16} and \emph{bfloat16} can work, either out of the box or in conjunction with techniques like loss scaling that help control the range of values in software. 

Our understanding of these issues needs to be deepened.  
To our knowledge, no study on the width of the range of values encountered during training has been performed and it has not been determined whether denormal numbers are necessary.
To fill this gap, we study here the distribution of tensor elements during training on public models like ResNet and BERT. 
The data show that the representation range offered by \emph{binary16} is fully utilized, with a high percentage of the values falling in the denormal range.  Unfortunately, on many machines performance suffers with every operation involving denormals, due to special encoding and handling of the range~\cite{subnormal, subnormal-old}.
This raises the question of whether they are needed, or are encountered much less often, when more than $5$ exponent bits are utilized.  
Indeed, \emph{bfloat16}, with its large representation range, does not support and does not appear to require denormal numbers; but it loses an order of magnitude in numerical accuracy vis-a-vis \emph{binary16}. Thus, we consider an alternative $1/6/9$ format, i.e., one having $6$ exponent bits and $9$ explicit mantissa bits, both with the denormal range and without it. 
We demonstrate that this format dramatically decreases the frequency of denormals, which will improve performance on some machines.

In Section~\ref{sec:related-work} we discuss $16$-bit formats that have been previously used in neural network training.
In Section~\ref{sec:fp-arith} we recall basic notions of floating-point arithmetic and define the notations we use.
Section~\ref{sec:experiments} presents our testing methodology and details the training settings for each tested network (Section~\ref{sec:network-details}). 
The results we gathered are detailed in Sections~\ref{sec:results} and, finally, we conclude in Section~\ref{sec:conclusion}.

\medskip
\section{Related work}
\label{sec:related-work}
\medskip

There have been several $16$-bit floating-point (FP) formats proposed for mixed-precision training of neural networks, including \emph{half}-precision~\cite{mp-train, DLS, DLS-2}, Google's \emph{bfloat16}~\cite{bfloat-1, bfloat-2}, IBM's \emph{DLFloat}~\cite{dlfloat} and Intel's \emph{Flexpoint}~\cite{flexpoint}, each with dedicated configurations for the exponent and mantissa bits.

Mixed-precision training using \emph{half}-precision was first explored by Baidu and Nvidia in~\cite{mp-train}.
Using \emph{half}-precision for all training data reduces final accuracy, because the formats fail to cover a wide enough representation range, i.e., small values in \emph{half}-precision might fall into the denormal range or even become zero.
To mitigate this issue, the authors of~\cite{mp-train} proposed a mixed-precision training scheme. Three techniques were introduced to maintain SOTA accuracy (comparable to \emph{single}-precision):
\begin{itemize}
    \item "master weights", i.e., a \emph{single}-precision copy of the weights, is maintained in memory and updated with the products of the unscaled \emph{half}-precision weight gradients and the learning rate. This ensures that small gradients for each minibatch are not discarded as rounding error during weight updates;
    \item during the forward and backward pass, matrix multiplications take \emph{half}-precision inputs and compute reductions using single-precision addition (using FMACS instructions as described in Section~\ref{sec:fp-arith}). For that activations and gradients are kept in \emph{half}-precision, halving the required storage and bandwidth, while weights are downcasted on the fly;
    \item "loss scaling", which empirically rescales the loss before backpropagation in order to prevent small activation gradients that would otherwise underflow in half-precision.  Weight gradients must then be unscaled by the same amount before performing weight updates.
\end{itemize}

Loss scale factors were initially chosen on a model to model basis, empirically. Automated methods like dynamic loss scaling (DLS)~\cite{DLS, DLS-2}, and adaptive loss scale~\cite{ALS} were later developed. These methods allow for the scale to be computed according to the gradient distribution.
Using the above techniques has become the standard workflow for mixed-precision training. It is reported to be $2$ to $11\times$ faster than \emph{single}-precision training~\cite{mp-train, DLS, DLS-2}, while also being able to achieve SOTA accuracy on widely used models.

Another mixed-precision format available in specialized hardware is Google's \emph{bfloat16}~\cite{bfloat-1}. It was first reported as a technique to accelerate large-batch training on TPUs (Tensor Processing Units)~\cite{tpu}, with minimal or no loss in model accuracy. 
The results presented in~\cite{bfloat-1} reported only the convolutional layers in the forward pass being computed in \emph{bfloat16}, leading to no accuracy loss on ResNet-50~\cite{resnet} trained on Imagenet~\cite{imagenet}.

In subsequent work~\cite{bfloat-2}, Intel and Facebook studied the effectiveness of \emph{bfloat16} on convolutional and recurrent neural networks, as well as on generative adversarial networks and industrial recommendation systems.
The experiments are based on simulations in which \emph{bfloat16} arithmetic is modeled by \emph{single}-precision operations whose results are then rounded to values that are also representable in \emph{bfloat16}.
The results suggest that \emph{bfloat16} requires no denormal support and no loss scaling, and therefore, decreases complexity in the context of mixed-precision training. Recently, Nvidia also made it available in its latest architecture, Ampere~\cite{ampere}.

IBM's \emph{DLFloat}~\cite{dlfloat} is a preliminary study of the $1/6/9$ format that we also examine.
\emph{DLFloat} was tested on convolutional neural networks, LSTMs~\cite{lstm} and transformers~\cite{transformer} of small sizes, implying the potential for minimal or even no accuracy loss. The studied format did not support a denormal range.
The authors briefly discuss the factors contributing to the hardware saving, without providing statistics on how they actually influence the training results.

\emph{Flexpoint}~\cite{flexpoint} is a blocked floating-point format proposed by Intel. It is different from the previously mentioned formats by that it uses $16$ bits for mantissa and shares a $5$-bit exponent across each layer.
\emph{Flexpoint} was proposed together with an exponent management algorithm, called Autoflex, aiming to predict tensor exponents based on gathered statistics from previous iterations. The format allowed for un-normalized values to be stored, which invalidated the possibility of a denormal range. For this reason we are not going to study it further in this work.

For all these studies, \emph{single}-precision training accuracy was achieved. But there is no mention of whether denormals, which significantly impact hardware area and compute speed, are encountered; neither is it tested whether or not they must be supported to achieve that accuracy.

\medskip

\medskip
\section{Floating-point arithmetic}
\label{sec:fp-arith}
\medskip

Formally defined in Def.\ref{def:fp}, a floating-point (FP) number is most often used in computing as an approximation of a real number. Given a fixed bit width for representation, it offers a trade-off between representation range and numerical accuracy. 

\medskip
\begin{defn}
\label{def:fp}
A real number $X$ is approximated in a machine by a nearby floating-point number:
\begin{equation}
x = M \cdot 2^{E-p+1},
\label{eqn:rep}
\end{equation}
where,
\begin{itemize}
    \item $E$ is the exponent, a signed integer $E_{min} \leq E \leq E_{max}$;
    \item $M \cdot 2^{-p+1}$ is the significand (sometimes improperly referred to as the mantissa), where $M$ is also a signed integer represented on $p$ bits.
\end{itemize}
\end{defn}
\medskip

From an engineering standpoint, the exponent is stored as an unsigned integer and offset from the actual value by the exponent bias. In the current work we only consider ``vanilla" bias, i.e., equal range around $0$, and for this reason we do not add a bias to our definition. For details on the implementation we refer the reader to the IEEE-$754$ standard~\cite{ieee}.

To avoid representation redundancy the finite nonzero FP numbers are \emph{normalized}. This is done by choosing the representation for which the exponent is minimum. However, this results in an abrupt convergence towards zero. For this reason, the IEEE-$754$ standard~\cite{ieee} defined the denormal range, that allows for gradual underflow towards $0$. 

The two different types of representations are as follows:
\begin{itemize}
    \item when the number is greater than or equal to $2^{E_{min}}$ and its representation satisfies $2^{p-1} \leq \abs{M_x} \leq 2^p-1$, we say that it is a \emph{normal} number;
    \item otherwise, one necessarily has $E=E_{min}$, and the significand adjusted according to that, with $\abs{M_x} \leq 2^{p-1}-1$; the corresponding FP number is called a \emph{subnormal / denormal} number.
\end{itemize}

From a numerical perspective, supporting denormal numbers offers a wider range of representation, which is desirable. However, traditional hardware implementation of denormals is generally slower, requiring more clock cycles.  Moreover, a denormal number is represented with fewer significant bits and with a resulting loss of precision. 

The normalized numbers are implemented using the implicit bit convention, in which a binary $1$ is appended to the most significant bit (MSB) of the significand, yielding a $p+1$ precision. That bit is also called a hidden bit since it isn't stored in the machine word. 
In such a system, some data (e.g. number $0$) cannot be expressed as a normal or denormal number. For this reason, in order to achieve a ``closed" FP system in which any operation is well specified, the standard allows for non-numeric data to be encoded as follows:
\begin{itemize}
    \item $E=0$ and $M=0$ is reserved to representing $\pm 0$;
    \item $E=E_{max}$ and $M= 0$ represents $\pm \infty$;
    \item $E=E_{max}$ and $M\neq 0$ represents NaN (Not a Number).
\end{itemize}

Considering all the implementation details, a formal definition of a FP system is given in Def.~\ref{def:fp-sys}. For more details on FP systems and their implementation we refer the reader to~\cite{ieee, handbook}.

\medskip
\begin{defn}
\label{def:fp-sys}
A floating-point number system is characterized by a quadruple:
\begin{equation}
s/e/p/d,
\label{eqn:sys}
\end{equation}
where,
\begin{itemize}
    \item $s$ is the number of sign bits ($1$ in every case considered here);
    \item $e$ the number of exponent bits;
    \item $p$ the number of mantissa bits
    \item $d$ is either the letter ``d" or ``n" depending on whether denormals are allowed or not.
\end{itemize}
\end{defn}


\medskip
\subsection{Mixed-precision training formats}
\label{sec:DL-formats}
\medskip

\begin{figure*}[htbp]
\centerline{\includegraphics[width=0.9\textwidth,keepaspectratio]{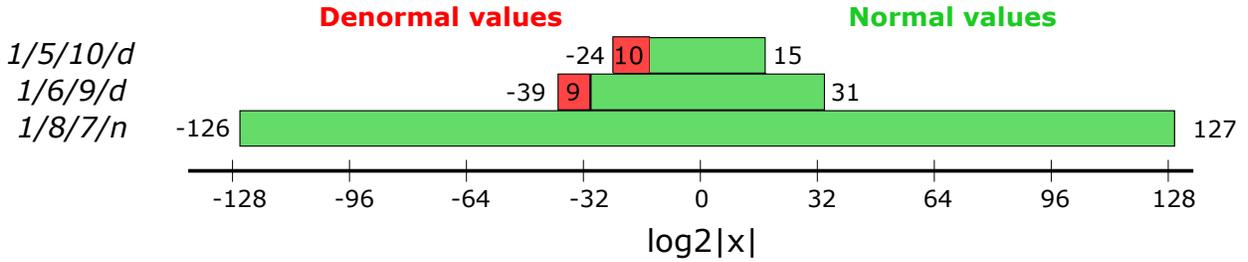}}
\caption{Visual representation of the representation range of $1/5/10/d$, $1/6/9/d$ and $1/8/7/n$, respectively.}
\label{fig:fp-formats}
\end{figure*}

As noted in Section~\ref{sec:related-work} the two $16$-bit mixed-precision formats already used in the deep learning community are \emph{half}-precision ($1/5/10/d$) and \emph{bfloat16} ($1/8/7/n$). To these we add our $1/6/9$ format. 
Following the notation introduced in Def.~\ref{def:fp-sys} we present the specific details in Table~\ref{tab:fp-formats} and a representation range visualization in Fig.\ref{fig:fp-formats}.

\begin{table}[htbp]
\caption{\vspace{1em} Characteristics of $16$-bit FP formats used for mixed-precision training. Following the notation of Def.~\ref{def:fp-sys}.}
\label{tab:fp-formats}
\begin{center}
\vspace{-.5em}
\begin{tabular}{|c||c|c|c|c|c|}
\hline
      & $e$ & $p$ & $E_{min}$ & $E_{max}$ & min. denormal \\
	\hline
	 $1/5/10/d$ & 5 & 10 & -14 & 15 & $\pm 2^{-24}$ \\
	\hline
	 $1/6/9/d$ & 6 & 9 & -30 & 31 & $\pm 2^{-39}$ \\
	\hline
	 $1/8/7/n$ & 8 & 7 & -126 & 127 & --- \\
	\hline
\end{tabular}
\end{center}
\end{table}

As observed in Fig.\ref{fig:fp-formats}, $1/8/7/n$ offers a very wide representation range, the same as IEEE-$754$ \emph{single}-precision. Thanks to its $8$-bit exponent, its normal range is already wider than the full range, including denormals, of $1/5/10/d$ or $1/6/9/d$, respectively. Also, it is public knowledge that Google TPU's~\cite{tpu} implementation does not offer any support for denormal numbers, and their models are able to converge without issues. This implies that $1/8/7/n$ mixed-precision training of neural networks does not rely on denormals to achieve SOTA accuracy and for this reason we are not going to discuss this format in what follows.

In comparison, $1/5/10/d$ offers a much narrower representation range, with only $5$ exponent bits. Although it has been adopted as the ``standard" mixed-precision in the community, this is largely due to its availability as part of the IEEE-$745$ standard. As we will show in Section~\ref{sec:results}, its full range is being used during training, which may lead to execution slowdown, depending on the hardware platform.
For this reason, in this work we introduce $1/6/9/d$, which by extending the exponent range by one bit, is able to almost completely avoid denormal numbers. 

\medskip
\subsection{Mixed-precision instructions}
\label{sec:DL-ops}
\medskip

Neural network training relies heavily on matrix multiplication, which is performed using multiply and accumulate operations. The IEEE-$754$ standard only defines instructions on one numerical format at a time, i.e., the input and output variables are represented in the same FP system. In the context of mixed-precision, operations that allow for higher precision outputs have become available.
The final result accuracy also depends on the number of rounding operations used, weather or not the instructions is fused. 
We introduce four possible instructions:
\begin{itemize}
    \item MAC: $a_{16} = \circ(a_{16} + \circ(x_{16} \cdot y_{16}))$;
    \item MACS: $a_{32} = \circ(a_{32} + \circ(x_{16} \cdot y_{16}))$;
    \item FMAC: $a_{16} = \circ(a_{16} + x_{16} \cdot y_{16})$;
    \item FMACS: $a_{32} = \circ(a_{32} + x_{16} \cdot y_{16})$,
\end{itemize}
where, $\circ$ represents any rounding operation performed and the subscript number gives the bit-width.
In the following sections we will discuss the effects of FMAC and FMACS on denormal frequency.

\medskip
\section{Experiments}
\label{sec:experiments}
\medskip

\subsection{Testing methodology}
\label{sec:testing}
\medskip

All experiments were carried out using the PyTorch framework with handwritten CUDA extensions to handle custom floating-point formats. 
In particular, we define a \emph{roundfp\_cuda(X, e, p, d)} kernel that takes a tensor $X$ of \emph{single}-precision FP values and rounds each element to the nearest-representable FP value with $e$ exponent bits and $p$ mantissa bits, where $e<=8$, $p<=23$ and $d$ specifies if the denormal range is allowed. 
We bind this kernel to a custom PyTorch function \emph{roundfp(X, e, p, d)} which performs rounding on both the forward and backward pass.

When defining a neural network model for our experiments, we insert \emph{roundfp} operations at the inputs and outputs of every layer, forcing all activations and activation gradients to be cast to the modeled FP format. The weight updates computation is kept in \emph{single}-precision.
The inner matrix multiplies (or convolutions) on both the forward and backward pass are performed using FMAC or FMACS instructions, meaning we do not round the products.

As a concrete example, when quantizing a \emph{torch.nn.Linear} layer with input $X$, weight $W$ and bias $B$, the forward pass is computed as:
$$ Y = R( R( R(W^T) \cdot R(X)) + R(B) ),$$
where $R(.)$ is the \emph{roundfp} function. Inside the matrix multiplication, $W^T \cdot X$, FMAC or FMACS instructions are used.

Accumulating sums of products using FMAC operations leads to significant loss in final accuracy due to reduced accumulator precision~\cite{mp-train}. For this reason, we only perform $8$ consecutive FMAC instructions before adding it to a \emph{single}-precision master accumulator, and resetting the $16$-bit accumulator to $0$.
Algorithm~\ref{algo:h-acc} details a fast and accurate dot product 
of half-precision vectors, returning a half-precision result, using this technique. We note here that the number $8$ of half-precision accumulations was empirically chosen; we did not investigate whether or not it can be greater than $8$ without affecting accuracy.
In what follows we will be using FMAC-$8$ to mean either this algorithm or a matrix multiplication or convolution in which such sums are accumulated in this manner.

\begin{algorithm}
\begin{algorithmic}[1]
    \STATE { $A^{(32)} = 0; a^{(16)} = 0$ }
    \FOR { $i \in [0, n)$ }
    \IF { $i \mod 8 = 0$ }
            \STATE { $A^{(32)} = A^{(32)} + a^{(16)}$ }
            \STATE { $a^{(16)} = 0$ }
        \ENDIF
        \STATE { $a^{(16)}$ = FMAC($a^{(16)}, w_i, x_i$) }
    \ENDFOR
    \STATE { $A^{(32)} = A^{(32)} + a^{(16)}$ }
    \RETURN {$R(A^{(32)})$}
\caption{FMAC-$8$. Dot product of half-precision input vectors $w$ and $x$ with half-precision output using FMAC instructions. Superscript number represents the bit width of the variable, $+$ denotes single-precision addition, and $R(.)$ is the \emph{roundfp} operation.}
\label{algo:h-acc}
\end{algorithmic}
\end{algorithm}

\medskip
\subsection{Training details}
\label{sec:network-details}
\medskip

We evaluated the importance of denormal numbers to neural network training on four different models.

\smallskip
\underline{ResNet}~\cite{resnet} is considered a classic convolutional neural network (CNN), primarily used for computer vision applications such as image classification and object detection. We trained two similar model depths, but using different datasets, with one significantly bigger than the other:
\begin{itemize}
    \item ResNet20 on Cifar10 dataset~\cite{cifar10}, trained for $250$ epochs with batch size $512$ and an initial learning rate $0.1$, decayed by $0.1\times$ at epochs $[100, 150, 200]$.
    \item ResNet18 on ImageNet dataset~\cite{imagenet} with $1000$ classes, trained for $90$ epochs with batch size $256$ and initial learning rate $0.1$, decayed by $0.1\times$ at epochs $[30, 60, 80]$.
\end{itemize}

\smallskip
\underline{BERT}~\cite{bert} is a state of the art transformer-based~\cite{transformer} model for natural language processing (NLP). Its training consists of two stages, pre-training and fine-tuning. For our scope we focused only on pre-training, since it is the more computationally demanding part. We studied two variations of the model:
\begin{itemize}
    \item BERT-Tiny~\cite{bert-tiny}, a shallow model with only two encoder layers;
    \item BERT-Base, one of the variants described in the original paper~\cite{bert}, with $12$ encoder layers.
\end{itemize}
Training was done using the OpenWebText dataset for $900k$ steps with sequence length $128$ and batch size $256$. We used a learning rate of $5e-5$ for BERT-Tiny and $1e-4$ for BERT-Base, respectively. The learning rate was scheduled to linearly ramp up at the beginning and then decay over the course of pre-training.

\smallskip
\underline{LSTM}~\cite{lstm} is a long short-term memory network with $2$ layers. Similar to the BERT model, we focused on a pretraining task. Training was done using the WikiText dataset with batch size $128$ and learning rate $1e-3$, for a total of $10k$ steps.

\smallskip
\underline{CVAE}~\cite{cvae} is a convolutional variational autoencoder model trained on a protein modeling task~\cite{anl}, using a private protein structure dataset.
The model was trained for $3k$ steps with batch size of $512$ and learning rate $1e-4$.

\medskip
\section{Results}
\label{sec:results}
\medskip

Matrix multiplications, at the core of fully connected and convolution layers, 
account for over 90 percent of the arithmetic in neural network training. For this reason, we monitor tensor elements in fully connected layers and convolutions. During simulations we log the denormal fraction of each layer's activations, weights, and activation gradients as they are computed and propagated through the network. From there we extract the highest denormal fraction observed across all of the tensors in the model. 
Most often, activation gradients have smallest magnitudes, since they become smaller with training\footnote{This is known as the vanishing gradient problem.}. This problem is partially solved by dynamic loss scaling (DLS). In addition, denormals may occur during the forward pass of training, as we will show.

We used each of the 16-bit floating-point formats in two training scenarios:
\begin{itemize}
    \item $16$-bit mixed-precision training without dynamic loss scaling (w/o DLS), i.e., none of the tensors are scaled in order to better fit the representation range;
    \item mixed-precision training with dynamic loss scaling (w/ DLS)~\cite{DLS}, which is a common practice in the community.
\end{itemize}

\medskip
\subsection{Using FMAC-8 accumulation}
\medskip

In Table~\ref{tab:denorm-ratio-h-acc} we document the maximum denormal fraction observed over the entire length of training on hardware that only offers FMAC instructions. For this we applied our FMAC-$8$ algorithm (Algo.~\ref{algo:h-acc}). This fraction can be very high for  $1/5/10/d$-precision. With DLS enabled, that fraction is reduced, but for LSTM it is still high, notably $60\%$ denormals in a single tensor at some point during training. This shows that denormals can be predominant in the forward pass as well, which is not scaled by the DLS technique. 
The situation is better with $1/6/9/d$-precision.   Without DLS there is strong reduction from the fractions seen with $1/5/9/d$, but for LSTM the fraction remains high.  Finally, the combination of $1/6/9/d$-precision and DLS effectively solves the problem of slowdown due to the prevalence of denormals.

\begin{table}[htbp]
\caption{\vspace{1em} Maximum denormal fraction over the entire course of training of ResNet20, BERT-Tiny, LSTM and CVAE using FMAC-8 accumulation. }
\label{tab:denorm-ratio-h-acc}
\begin{center}
\begin{tabular}{|c||c|c|c|c|}
\hline
     \multirow{3}{3em}{Model} & \multicolumn{4}{c|}{maximum denormal ratio}  \\
    \cline{2-5}
        & \multirow{2}{4em}{$1/5/10/d$} & $1/5/10/d$ & \multirow{2}{4em}{$1/6/9/d$} & $1/6/9/d$ \\
        & & w/ DLS & & w/ DLS \\
	\hline
	 ResNet20 & $0.95$ & $0.05$ & $0.25$ & $0$ \\
	\hline
	 BERT-Tiny & $0.9$ & $0.25$ & $0.1$ & $0.0005$ \\
	\hline
	 LSTM & $0.9$ & $0.6$ & $0.6$ & $0.001$ \\
	\hline
	 CVAE & $0.8$ & $0.15$ & $0.3$ & $0.1$ \\
	\hline
\end{tabular}
\end{center}
\end{table}

\begin{figure*}[htbp]
\centerline{\includegraphics[width=0.9\textwidth,keepaspectratio]{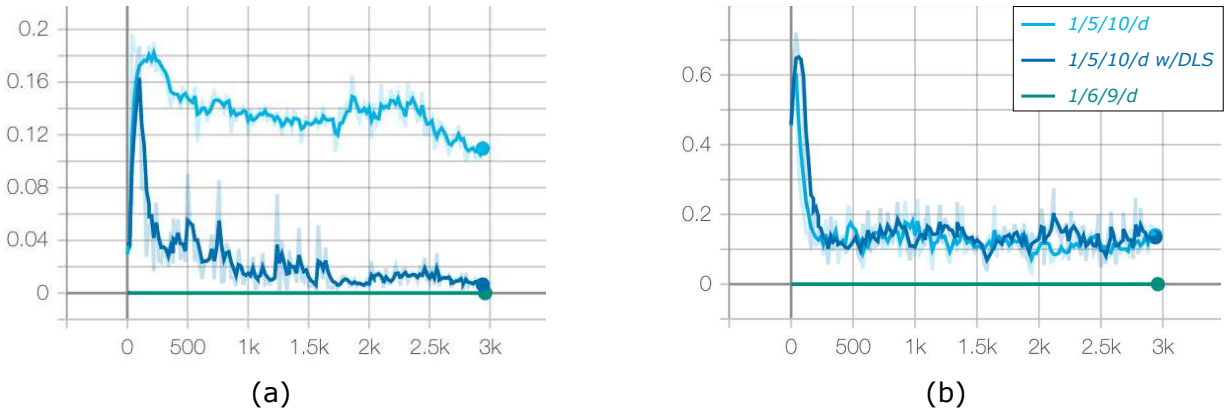}}
\caption{Tensorboard visualization of CVAE training.  The curves show the denormal fraction during backward propagation over $3k$ steps. Values logged from (a) $4^{th}$ convolution layer and (b) $2^{nd}$ deconvolution layer.
The three runs utilized $1/5/10/d$, $1/5/10/d$ with DLS, and $1/6/9/d$ (in green, at the bottom).
Denormals occur frequently, with or without DLS, when using the $1/5/10/d$ format.}
\label{fig:denorm-percentage}
\end{figure*}

Fig.\ref{fig:denorm-percentage} shows tensorboard screenshots from our CVAE runs, that showcase different behaviour for different tensors.
Taking a closer look at one of the convolution layers (Fig.~\ref{fig:denorm-percentage}(a)), we see that it is more than $80\%$ denormal for $1/5/10/d$ training. Applying DLS reduces the fraction to $10\%$. 
During the same run, one of the deconvolution layers (showcased in Fig.~\ref{fig:denorm-percentage}(b)) does not exhibit the same reduction when using DLS, the denormal ratio being approximately the same for both $1/5/10/d$ runs (w/ and w/o DLS).
For these particular layers, $1/6/9/d$ is able to fully represent the values using only its normal range.

\medskip
\subsection{Using FMACS accumulation}
\medskip

Some specialized hardware architectures offer instructions such as FMACS, which allows for accumulation to happen in higher precision. Empirically, this avoids any denormal intermediate computation.
This instruction can be found in Nvidia's GPU architecture, based on TensorCores~\cite{tensorcores}.
When testing this setting we were able to take advantage of GPU architecture and gained speed up, which allowed us to collect statistics on even bigger models like BERT-Base and ResNet18-ImageNet.

In Table~\ref{tab:denorm-ratio-bc}, one can observe that using single precision additions in the matrix multiplies reduces the maximum denormal fraction for some networks, like BERT-Tiny. The reason likely is that the threshold for denormals is much lower in single precision.  But for CVAE, the use of FMACS instructions doesn't reduce denormals much, without DSL, and not at all with DLS. 
The data do suggest that when FMACS is used in conjunction with the $1/6/9/d$ format, the use of DLS is not needed to make the denormal fraction negligible or at least acceptable.
(Note that we cannot compare the ResNet20 with the ResNet18 results, 
because the datasets used are different, with Imagenet being much larger.)

\begin{table}[htbp]
\caption{\vspace{1em} Maximum denormal ratio over the entire course of training of ResNet18, BERT-Tiny, BERT-Base, LSTM and CVAE on hardware that supports FMACS instructions. }
\label{tab:denorm-ratio-bc}
\begin{center}
\begin{tabular}{|c||c|c|c|c|}
\hline
     \multirow{3}{3em}{Model} & \multicolumn{4}{c|}{maximum denormal ratio}  \\
    \cline{2-5}
        & \multirow{2}{4em}{$1/5/10/d$} & $1/5/10/d$ & \multirow{2}{4em}{$1/6/9/d$} & $1/6/9/d$ \\
        & & w/ DLS & & w/ DLS \\
	\hline
	 ResNet18 & $0.98$ & $0.36$ & $0.5$ & $0.003$ \\
	\hline
	 BERT-Tiny & $0.6$ & $0.25$ & $0.009$ & $0.0003$ \\
	\hline
	 BERT-Base & --- & $0.2$ & --- & $0.003$ \\
	\hline
	 LSTM & $0.3$ & $0.23$ & $0.05$ & $0.004$ \\
	\hline
	 CVAE & $0.7$ & $0.15$ & $0.2$ & $0.1$ \\
	\hline
\end{tabular}
\end{center}
\end{table}

\medskip
\subsection{Training with flushed formats}
\medskip

The data we have presented thus far show that using the $1/6/9/d$ format reduces the frequency of denormal values enough that it can speed up training on hardware that suffers a slowdown whenever a denormal value is created.
Guided by the very low denormal ratios reported on $1/6/9/d$ with DLS, we also trained using an $s/e/p/n$ FP system (Def.~\ref{def:fp-sys}), in which denormals are flushed to $0$, with no slowdown.   The question then is whether training speed or accuracy suffers.

The convergence results are detailed in Table~\ref{tab:convergence}. All the networks achieved SOTA accuracy using our proposed format, $1/6/9/n$ with DLS. The same cannot be stated about training with $1/5/10/n$, where we observed a significant decrease in final model accuracy\footnote{No hyperparameter tuning was performed for these tests.}.   Note also that the smallest tested models, which are ResNet20, LSTM and CVAE, 
converge in $1/6/9/n$ without DLS.

\begin{table}[htbp]
\caption{\vspace{1em} Convergence results for training with a $s/e/m/n$ scheme, with and without DLS. \cmark~indicates SOTA convergence, \xmark~indicates fail to converge and $\mathsf{D}$ indicates a degradation in the final loss.
}
\label{tab:convergence}
\begin{center}
\begin{tabular}{|c||c|c|c|c|}
\hline
    \multirow{2}{3em}{Model} & \multirow{2}{4em}{$1/5/10/n$} & $1/5/10/n$ & \multirow{2}{4em}{$1/6/9/n$} & $1/6/9/n$ \\
        & & DLS & & DLS \\
	\hline
	 ResNet20 & \xmark & $\mathsf{D}$ & \cmark & \cmark \\
	\hline
	 ResNet18 & \xmark & $\mathsf{D}$ & $\mathsf{D}$ & \cmark \\
	\hline
	 BERT-Tiny & \xmark & \xmark & $\mathsf{D}$ & \cmark \\
	\hline
	 BERT-Base & \xmark & \xmark & \xmark & \cmark \\
	\hline
	 LSTM & \xmark & \cmark & \cmark & \cmark \\
	\hline
	 CVAE & \xmark & $\mathsf{D}$ & \cmark & \cmark \\
	\hline
\end{tabular}
\end{center}
\end{table}

\begin{figure*}[htbp]
	\centering
	\includegraphics[width=0.9\linewidth]{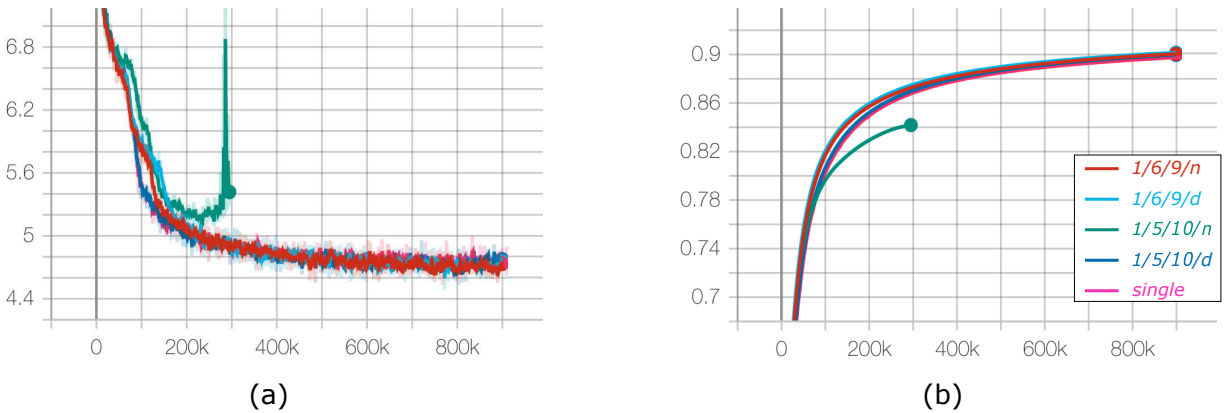}
\caption{Tensorboard visualization of (a) training loss curve and (b) next sentence prediction accuracy for BERT-Tiny mixed-precision training with DLS and $1/6/9/n$, $1/6/9/d$, $1/5/10/n$ and $1/5/10/d$, respectively. \emph{Single}-precision training without DLS is also included. Note that \emph{single}-precision curves are underlying. }
\label{fig:BERT-train}
\end{figure*}

Fig.\ref{fig:BERT-train} contains tensorboard screenshots for training BERT-Tiny in mixed-precision with DLS vs. \emph{single}-precision. For reference we include the \emph{single}-precision training curves, despite that they are underlying and barely visible.
The loss curves in Fig.\ref{fig:BERT-train}(a) show that $1/5/10/n$ is the only run that diverges after $200k$ training steps, while all others converge as expected and achieve SOTA final accuracy. This can also be observed on the Next Sentence Prediction (NSP) training accuracy in Fig.\ref{fig:BERT-train}(b).

\medskip
\section{Conclusions}
\label{sec:conclusion}
\medskip

We studied the occurrence of denormal numbers during mixed-precision training of neural networks. We analyzed two $16$-bit formats: $1/5/10/d$, which is the same as the IEEE-$754$ \emph{half}-precision and a novel $1/6/9/d$ format that has one more exponent bit and one fewer significand bit.

Our data suggests that the $1/6/9$ format suppress most denormals because it significantly expands the representation range of normalized numbers.  This can speed up training on many machines, those that slow down on any computation involving denormals.
Despite the loss of one bit in the signifcand, we observed no loss of training speed or final accuracy compared with $1/5/10$, or for that matter with \emph{single}-precision ($1/8/23$) training.

We observed other advantages. No re-tuning of hyperparameters like the batch size or learning rate was needed.

We saw that the $1/5/10/d$ format can achieve SOTA as well, though $1/5/10/n$ cannot. 
We did not see any difference in number of epochs to accuracy between $1/6/9$ (``d" or ``n") and $1/5/10/d$.
Thus, denormals are essential when using $1/5/10$, even when DLS is employed (see Table~\ref{tab:convergence}).
And they occur frequently enough with $1/5/10$ to be a performance problem on many platforms.

We found that with DLS, $1/6/9/n$ trains as accurately and as fast as \emph{single}-precision, does not suffer a slowdown due to denormals. And that $1/6/9/d$ is fast and reliable with or without DLS in most cases.
Furthermore, we saw that a limited amount of \emph{half}-precision accumulation is allowable without changing these conclusions.

\bibliographystyle{IEEEtran}
\bibliography{ref}

\vspace{12pt}
\color{red}

\end{document}